# Deep Learning on Mobile Devices – A Review

Yunbin Deng
FAST Labs, BAE Systems, Inc. Burlington MA 01803

**ABSTRACT**

Recent breakthroughs in deep learning and artificial intelligence technologies have enabled numerous mobile applications. While traditional computation paradigms rely on mobile sensing and cloud computing, deep learning implemented on mobile devices provides several advantages. These advantages include low communication bandwidth, small cloud computing resource cost, quick response time, and improved data privacy. Research and development of deep learning on mobile and embedded devices has recently attracted much attention. This paper provides a timely review of this fast-paced field to give the researcher, engineer, practitioner, and graduate student a quick grasp on the recent advancements of deep learning on mobile devices. In this paper, we discuss hardware architectures for mobile deep learning, including Field Programmable Gate Arrays (FPGA), Application Specific Integrated Circuit (ASIC), and recent mobile Graphic Processing Units (GPUs). We present Size, Weight, Area and Power (SWAP) considerations and their relation to algorithm optimizations, such as quantization, pruning, compression, and approximations that simplify computation while retaining performance accuracy. We cover existing systems and give a state-of-the-industry review of TensorFlow, MXNet, Mobile AI Compute Engine (MACE), and Paddle-mobile deep learning platform. We discuss resources for mobile deep learning practitioners, including tools, libraries, models, and performance benchmarks. We present applications of various mobile sensing modalities to industries, ranging from robotics, healthcare and multi-media, biometrics to autonomous drive and defense. We address the key deep learning challenges to overcome, including low quality data, and small training/adaptation data sets. In addition, the review provides numerous citations and links to existing code bases implementing various technologies. These resources lower the user's barrier to entry into the field of mobile deep learning.

## 1. INTRODUCTION

Deep learning is the key enabler for many recent advances in artificial intelligence applications. The greater benefit will come when the artificial intelligence technologies becomes ubiquitous in mobile applications, such as automatous driving, affordable robots for home, and more intelligent personal assistance on mobile phone. Compared with the traditional mobile sensing and cloud computing paradigm, the advantages of deep learning and inference on the mobile device are four folds: 1) The saving of communication bandwidth. The more computing done on the mobile device, the less data needs to be sent to the cloud. 2) The reduction of cloud computing resource cost. The cost of maintaining or even renting cloud computing resources can be prohibitive for some applications. Computation on mobile devices becomes possible as mobile devices become more computationally powerful. 3) The quick response time. If all computation can be performed locally, there will be no overhead in communication time or any concerns of server reliability. For some applications, such as in the health care and military domains, this response time is very critical. 4) Mobile computing keeps the sensory data on the local device, thus greatly improving user data privacy. This is particularly true for home robotics applications.

Research in deep learning for mobile and embedded devices has thus becomes a very hot topic. The subtopics cover hardware architecture considerations, algorithmic optimizations, and choices for mobile deep learning platforms. As the research and development progresses, many tools, data, and models are made publicly available. For a beginner to mobile deep learning, the amount of information available in this topic can be huge. This paper gives the reader a solid grasp on these topics enabling them to quickly start to make contributions to this fast growing field. This is achieved by reviewing and comparing existing hardware, software, algorithmic, and platform choices, listing available resources, guidance in making mobile deep learning applications, and discussing the challenging problems to be solved. This paper reviews recent scientific and technological advances to bring deep learning to mobile and embedded devices. However, it does not cover the basics of machine learning (ML) and deep learning. Readers should refer to more basic tutorials in deep learning if necessary, e.g. [29][30].

This paper begins with the hardware aspects for deep learning. The hardware architecture design tradeoffs for system accuracy, power consumption, and cost will be discussed in Section 2. A state-of-the-industry review will be given in Section 3 to detail some of the most promising mobile deep learning computation platforms and libraries. The pros, cons, and application domains of each platform will be discussed. It details potential development paths based on the timeframe available to the developer. Different paths are available given development timeframes from hours, days, weeks, to months. The software and algorithmic advancement for mobile deep learning are described in Section 4. Optimizations of deep learning algorithms, such as quantization, pruning, compression, and approximation have to be made while retaining as much performance accuracy as possible. For the convenience of the mobile deep learning practitioners, Section 5 provides a list of tools, models, and some publicly available codebases to start with. These resource significantly lower the barriers of entry for beginners. For readers interested in developing real mobile deep learning applications, Section 6 showcases some recent successes in mobile deep learning development in various application domains, ranging from robotics, autonomous driving, healthcare, multi-media, biometrics, entertainment and defense. Although a great deal of resources and progress has been made for mobile deep learning, challenges remain. Section 7 outlines a few key challenges, which could serve as important research topics for graduate students.

## 2. HARDWARE ARCHITECTURES FOR MOBILE DEEP LEARNING

Advances in deep learning are mostly attributed to the availability of powerful computation resources and large amounts of training data. One of the key contributors to increased computation power comes from low cost GPUs. A high performance deep learning application requires a high quality machine learning model trained with a large amount of data, often on the order of terabytes. This is typically achieved using high performance Central CPUs, GPUs, Tensor Processing Units (TPUs), or their combination. Once the model is built, it can be deployed on a mobile device with much less computation power, including CPU, ASIC, FPGAs, and mobile GPUs. For mobile deep learning applications, the computation hardware architecture decisions are based on the computation performance, system SWAP, and cost (in term of time and money for the development) considerations.

### 2.1 CPU

Although CPUs can be cumbersome for building state-of-the-art deep learning models when the training data is very large, they are a valid option for training or adapting models to small amounts of data and to deploy pre-built deep learning models. The obvious advantage of deploying mobile deep learning on a CPU is that CPUs are available for most tablets and mobile phones. In addition, these devices often come with powerful CPUs and have a large selection of built-in sensors. This means there is no extra hardware cost to enable more AI applications for a potential huge market. Real-time applications can be realized on a standard CPU with little or no extra effort. However, a full grasp of a few important techniques can be very beneficial to building computation- and power-efficient applications for CPUs.

- Use the highly optimized linear algebra library LAPACK/BLAS, most existing deep learning frameworks already take this into account [1];
- Take full advantage of multi-core multi-threaded CPUs for computation parallelization;
- Make use of CPU manufacturers-specific mathematic library;
- Conduct tradeoff studies of model size, numeric precision, computation speed, and system performance.

Many successful deep learning applications have been implemented on CPUs, including voice, fingerprint, and face biometrics on mobile devices. The market has driven the mobile technology to a point where the CPUs on mainstream mobile devices are just as powerful as CPUs on mainstream tablets and laptops.

## 2.2 GPUs and Mobile GPUs

The GPU is a highly parallel processor by physical design. It typically contains thousands of repetitive small cores, making it much more efficient to do certain types of repetitive computations on matrices than general purpose CPUs. The GPU was originally designed mostly for graphics and gaming, but it became the key enabler for building high performance machine learning models in the past few years. Now, GPUs are widely used in embedded systems, mobile phones, PCs, workstations, and game consoles. NVIDIA offers a wide range of GPU products. The NVIDIA GPUs are very useful for machine learning engineers, as they provides a powerful programming platform and application programming interface (API) called Compute Unified Device Architecture (CUDA) [3]. CUDA works with a few of the most popular programming languages, such as Python, C, C++, and FORTRAN. For example, pyCUDA makes it much easier for a Python programmer to utilize the GPU resources. Some of the most popular NVIDIA GPUs relevant to deep learning are listed in Table 1. The performance is measured in terms of double precision TFLOPS, the number of tera floating point operation per second. For the serious deep learning researcher, a GPU with performance at the level of GTX 1080 Ti or higher should be chosen to build state of the art models when large amounts of training data are available. This GPU provides 332 TFLOPS computation power. Many supercomputers are made of Tesla GPUs, while the Drive PX GPUs are popular system-on-a-chip component for autonomous driving [31].

**Table 1: The performance of some NVIDIA deep learning GPUs**

| nVidia GPU | GeForce GTX | Tesla | Drive PX |
| --- | --- | --- | --- |
| Processing Power (TFLOPS) | 10s ~ 100s | 100s ~ 1000s | 10s ~ 100s |
| Application | Gaming, ML | AI Cloud, Supercomputer | Autonomous driving |

NVIDIA provides mobile Geforce and Quadro GPU's to laptop manufacturers. There are many mobile GPUs available from different vendors. A comprehensive list of graphics cards for notebooks and smartphones can be found in [32]. A list of GPUs made for androids is listed in [33]. One of the most notable smartphone system on a chip (SoC) is the Qualcomm Snapdragon chip [34]. The SoC features multi-core CPU, a GPU, GPS, wireless modem, among others, making many applications of deep learning on a smartphone feasible.

## 2.3 FPGA

While the CPU is made for extremely general purpose computation, and an ASIC is almost made exclusively for a specific application, the FPGA lies in between. FPGAs can be (re)programmed by 'firmware' to perform many specific applications very efficiently. It also lies in between in terms of system development time and power consumption. At the basic level, FPGAs use flip-flop circuits to implement sequential logic functions and table look-ups, i.e. memory circuit to implement combinational logic function. The logic functions are implemented by programming memory, which also controls switch circuit connections, thus FPGAs do not need to explicitly perform the logic operation once it is programmed. Modern FPGAs tend to take a SoC approach to integrate process core, communication core, and memory on a single chip.

FPGA venders, such as Xilinx and Altera, have made many software tools to make FPGA programming easier. While traditional FPGA programmers need to grasp knowledge of digital circuits and hardware description language (HDL), the trend of FPGA tools are towards high-level synthesis of circuits (HLS) [40]. The HDL software tools do not require the developer to perform complex low-level hardware operation control. There are five main categories of HLS tools, but the parallel computing framework OpenCL is most relevant for mobile deep learning. OpenCL is a C-based language and is an open source, standardized framework for algorithm acceleration. Programs written in OpenCL can be executed on GPUs, DSPs and FPGAs. OpenCL can be considered as an open source, royalty-free version of CUDA. Example implementations of Convolutional Neural Networks (CNN) and Deep Neural Networks (DNNs) on FPGA can be found here [36][37].

## 2.4 ASIC and TPU

The ASIC goes one step further than the FPGA to design and fabricate specific chips for an application. It has the potential to achieve even better SWAP, but at the cost of long development time. Building ASICs for mobile machine learning and artificial intelligence has a very long history. It was called neuromorphic engineering in the early days [38][39]. Many analog, digital, mixed-signal, memristor, and spike circuits are designed to mimic auditory, vision, and brain functions [41][42].

There are many ASICs developed for AI applications. TPU is one developed specifically for neural network machine learning by Google. It was used in the recent AlphaGo vs Lee Sedol man-machine Go games. Although not commercially available, TPU is offered to the machine learning researcher for free trial and is accessible through its cloud computing service. TPU provides 10s~100s TFLOPS and can fit into the rack slot of common cloud computer servers.

## 2.5 Comparison of CPU, GPU, FPGA, and ASIC for Mobile Deep Learning

As mentioned in previous subsections, the CPU and GPU are general purpose computing platforms, thus offering the most flexibility. Early algorithmic performance study of a deep learning application should make use of CPUs and GPUs to get a preliminary idea of achievable performance. CPUs and GPUs support full precision computing and can afford computation intensive models, which often means higher prediction accuracy is achievable. However, GPUs and CPUs are less power efficient. ASICs can be much more energy efficient as the hardware is made specifically for a certain computation. Yet, the design and development of ASIC chips can be very time consuming. An ASIC is thus only used when the algorithm research is settled and the system has a very lower power budget. FPGAs offers a tradeoff among power consumption, prediction accuracy, and system development speed. The typical achievable power consumption and ML accuracy metrics for these four computation architectures are illustrated in Figure 1 [27].

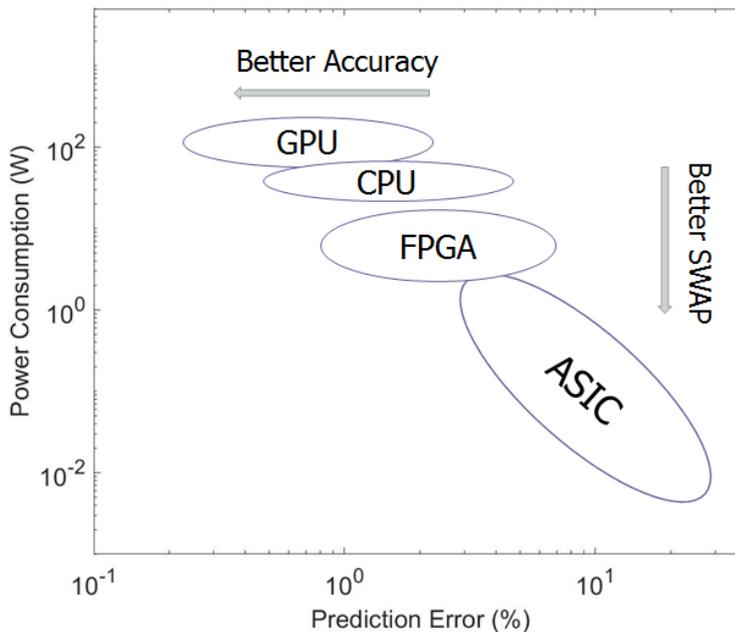

**Figure 1. The rough operation performance region of the four hardware architectures.**

## 3. MOBILE DEEP LEARNING PLATFORMS AND LIBRARIES

In the past few years, many platforms have been developed by the major internet companies for deep learning application on mobile devices. Depending on time to market constraints, different platforms should be chosen. Choosing the right platform can make a significant difference in development time, cost, and final system performance. The following subsection describes approaches with development time ranging from hours to years.

### 3.1 Building Mobile Deep Learning Solutions Leveraging Cloud AI services

If deep learning on the cloud is appropriate for the application, the quickest way to deploy deep learning capability on a mobile device is to take advantage of many existing cloud artificial intelligence APIs.   In this case, the mobile device serves as a sensor and user interface. These APIs offer ready-to-use AI capabilities in machine learning, speech recognition, computer vision, natural language processing, AI assistant, knowledge discovery, personality and emotion analysis, and search. Some of these well-known services include Microsoft Cognitive Services, Google Cloud Vision, IBM Watson Services, Amazon Rekognition, and Baidu EZDL.  If the application to be developed can use existing well-tested APIs, than this use can dramatically reduce development cost.

For developers with sufficient experience with existing cloud AI service APIs, the development timeframe can be as short as hours or days. However, it may be that existing APIs may not offer the desired capability, or performance may not be fully satisfactory on the data.  In that case, deeper investigation is needed to leverage lower level existing tools, models, data, detailed in the following subsection.

### 3.2 Building Mobile Deep Learning Solutions Leveraging Existing Platforms

Some deep learning applications, or at least the inference part, needs to be implemented on mobile devices without relying on cloud machine learning.  Many deep learning platforms can still be leveraged for mobile applications. A few of the most notable mobile deep learning platforms are list below.
- TensorFlow Lite from Google: TensorFlow is an open source tool developed by the Google Brain team for deep learning model training and deployment.  Its flexible architecture allows deployment to a variety of platforms from CPU, GPU, TPU, to mobile and edge devices. The use of the Python language makes it popular among developers. There is an easy pipeline to bring TensorFlow models to mobile devices and open source examples are provided to deploy TensorFlow to Android devices. For beginners, Keras is an easy python API to make use of TensorFlow to build models and run evaluation. TensorFlow Lite is the solution for running TensorFlow models on mobile and embedded devices. It was optimized for accuracy, low latency, small model size, and portability to both Android, iOS, and other internet of things (IoT) devices. The TensorFlow tools are available at https://www.TensorFlow.org/
- Caffe2 from Facebook: Caffe2 is a lightweight, modular, and scalable deep learning framework. It offers cross-platform libraries for deployment on the cloud or mobile devices. Caffe models can be under 1MB of binary size and are built for speed. The support ranges from ARM CPU, iPhone GPU, to Android GPU. In addition, it has support for importing models trained using CNTK and PyTorch. The Caffe2 is available at https://caffe2.ai/
- Core ML for IOS: core machine learning was developed by Apple Inc. for IOS 11. It was built on top of low-level primitives: accelerates basic neural network subroutines (BNNS), and metal performance shaders (MPS) to automatically minimize memory footprint and power consumption.  It has direct support for Keras, Caffe, Scikit-learn, XGGoost, and LibSVM machine learning tools [Keras, Caffe, scikit, XGGoost, Libsvm]. In addition, models built using Caffe and TensorFlow model can be converted to CoreML model in a few lines of code. Online documentation is available at https://developer.apple.com/documentation/coreml
- Snapdragon neural processing SDK from Qualcomm: It is designed to run DNN models trained in Caffe, Caffe2, ONNX, or TensorFlow on Qualcomm Snapdragon mobile platforms. It automatically identifies the best target core for inference, weather that is CPU, GPU or DSP.  Today, Snapdragon chips exist on about half of android phones. [https://developer.qualcomm.com/software/qualcomm-neural-processing-sdk
- DeepLearningKit for Apple Devices:  DeepLearningKit is a DL framework for tvOs (TV), iOS (iPhone and iPad), and OS X (MacBook and iMac). It supports CNN models trained in Caffe. It runs on mobile GPU. It is fast, but is unmaintained since 2016. The DeepLearningKit is available on github https://github.com/DeepLearningKit/DeepLearningKit
- Mobile AI Compute Engine (MACE): MACE is a deep learning inference framework optimized for mobile heterogeneous computing platforms. It supports only not a wide range of model formation, such as TensorFlow, Caffe, and ONNx, but also has techniques to protect the model, such as converting models to C++ code. It is optimizes run time, memory usage, library footprint, and UI responsiveness. The  MACE tool is available at https://github.com/XiaoMi/mace

- Paddle-Mobile: Paddle-Mobile is part of a bigger deep learning project called PaddlePaddle and focuses on embedded platforms. It support ARM CPU, Mali GPU, Android GPU/GPU, iOS, FPGA-based development boards, Raspberry Pi, and other ARM-Linux development boards. The Paddle-Mobile is available on https://github.com/PaddlePaddle/paddle-mobile

**Table 2. Comparison of existing platforms and their supported, mobile OS, DNN architecture, computing device, and model format.**

| ML platform | Mobile OS | DNN Architecture | GPU/CPU | Model Format |
|---|---|---|---|---|
| TensorFlow / TFLite | iOS, Android | Most DNNs | Yes / Yes | TensorFlow |
| Caffe2 | iOS, Android | CNN | Yes / Yes | Caffe2, CNTK, PyTorch |
| CoreML | iOS | CNN, RNN, SciKit | Yes / Yes | Keras, TensorFlow |
| Snapdragon | Android | CNN, RNN, LSTM | Yes / Yes | Caffe, Caffe2, TensorFlow |
| DeepLearningKit | iOS | CNN | Yes / Yes | Caffe |

### 3.3 Mobile deep learning libraries

#### 3.3.1 Keras

Some deep machine learning toolkits, such as TensorFlow, require the user to have very good knowledge of both machine learning and software engineering. Users with both skillsets are very limited, thus the adoption of ML was slow. The idea of Keras is to greatly simplify the learning and deployment of DL. By designing an easy to use high-level API and hide all unnecessary machine learning and software details, Keras python enables users to use TensorFlow, CNTK, or Theano without going through the learning curve to study each underlying library in detail. Keras code and documentation are available at https://keras.io/. The Keras project is so successful that it is fully integrated into TensorFlow now. Keras is one of the most popular tool for beginners and researchers in deep learning. There are numerous open source deep learning projects implemented in Keras available on github.

#### 3.3.2 Torch/PyTorch

PyTorch is a Python open source ML library based on Torch. Torch itself is an open source tool for machine learning and scientific computing based on Lua programming language. Although Torch is no longer in active development, PyTorch is very popular among academic researchers. There are also many open source projects and AI applications designed based on PyTorch. Pytorch is available at https://pytorch.org/.

#### 3.3.3 MXNet

MXNet from Apache Software Foundation. MXNet is a flexible and efficient library for deep learning. It packs all code in a single source file. The advantage is cross platform (iOS and android), easy porting, and usable in most major programming languages. However, only CPU is supported for deployment. MXNet is available at https://mxnet.incubator.apache.org.

#### 3.3.4 CNNDroid

CNNDroid is an open source library for speed up execution of CNN on Android devices. The GPU accelerated CNNs for Android Support Caffe, Torch, and Theano models and achieves 30~40 times speedup using GPU model vs CPU on AlexNet. CNNDroid is available at https://github.com/ENCP/CNNdroid.

### 3.3.5 JaveScript Libraries

JaveScript is one of the core programming language for web-based application. There are quite a few JavaScript libraries designed for deep learning development entirely in a web browser, including ConvNetJS, DeepLern.js, Keras.js, TensorFlow.js, Brain.js, and others. There is no need to install many libraries or drivers. These libraries are very convenient for building mobile deep learning demos. This also makes mobile machine learning more interactive. Some of them are for DNN inference only, but others are capable of DNNs training as well. Some are capable of training and inference using CPU, such as ConvNetJs. Others can make use of GPU for training and inference, such as DeepLearn.js. Keras.js runs the Keras model in browser with GPU support.

For developers with sufficient experience with existing platforms and libraries, the development of a mobile deep learning prototype can take from a few weeks or a few months. The next step is to test the application in the field. The real tests may find the computation, storage, memory footprint, or SWAP aspects need to be further improved. In particular, as the system gathers more domain specific data, you may be able to improve the deep learning system performance. The ML designer needs to have even deeper knowledge of deep learning algorithms and software to push its performance to a higher level. The next section summarizes some progress in mobile deep learning algorithms to further improve performance in these aspects.

### 3.4 Conversion of models trained on different platforms and libraries

Each set of deep machine learning libraries and tools has its own advantages, disadvantages, and use cases. The models trained using these tools all have their own format. To make the best use of all these tools and libraries, it may be beneficial to be able to convert the model format from one to the other. Thus there are a few projects developed for this purpose. One of the most powerful tools for model conversion is MMdnn. This tool can convert TensorFlow, CNTK, Keras, Caffe, PyTorch, MXNet, and CoreML formats model into an intermediate representation, which can then be converted into one of these native target format [28].

## 4. OPTIMIZATION FOR MOBILE DEEP LEARNING

Efforts have been taken in the domain of software, algorithmic, and joint software and hardware co-design to optimize the SWAP performance of mobile deep learning applications. This section describes these efforts.

### 4.1 DeepX Software Accelerator

One of the key requirements of deep learning on mobile devices is lower resource requirements, including memory, storage, computation, and power consumption. In addition to the progress on hardware design, there is also progress made on software approaches. DeepX is such a software accelerator designed for low power deep learning inference on mobile devices [43]. It consists of two resource control algorithms for deep learning inference. One algorithm decomposes the DNN architecture into blocks of various types, which are more efficiently executed using heterogeneous local GPU and CPU processors. The other algorithm performs resource scaling to adjust DNN model architecture, thus the overhead of each block.

### 4.2 Algorithmic optimizations for SWAP

The original motivation for going deeper in neural networks for machine learning is for better representation learning. These learned representations capture complex non-linear embedding in the feature space and grant the model better accuracy and better generalization performance. These models are often deep, contain millions of parameters, and often not ideal for mobile deep learning deployment. Recently, some very interesting work was done to make these high accuracy models more suitable for mobile applications with little or no sacrifice in accuracy. This section summarizes some of these ideas, including quantization, pruning, compression, and approximations of DNNs.

#### 4.2.1 CNN Architecture Optimization

Recent ResNet enabled training of DNN with over hundreds of layers achieved impressive performance in accuracy. However, the accuracy performance does not scale linearly with the number of layers. For mobile applications, a good trade off can be made to pick the architecture to balance accuracy and other SWAP considerations. The SqueezeNet work of Iandola proposed to optimize CNN for fewer parameters using three strategies [4]:

- Replace majority of the 3x3 filters with 1x1 filter.
- Decrease the number of input channels to the filters
- Down sample in later layers in the network

The SqueezeNet is able to achieve AlexNet level accuracy on ImageNet task with 50 times few parameters. With additional model compression techniques, detailed in subsections below, the SqueezeNet achieved 0.5MB size, or over 500 times smaller than AlexNet.

### 4.2.2 Pruning, Quantization, and Huffman Coding

The "deep compression" work by Han used techniques of pruning, quantization and Huffman coding to reduce model size by a factor of 35~49 without affecting accuracy [44]. The pruning basically removes all small-weight connections. Typically, the weight parameters can be reduced by a factor of 10 using pruning. The next steps are quantization and weight sharing. The idea is to cluster all the weights with similar values together and store these weights in a codebook. The last step applies the well-known Huffman coding to assign shorter code to more common symbols.

It is also a common practice to reduce precision from 32 bits to 16 bits or less for mobile applications. Different approximation methods, such as fixed point approximation, dynamic fixed point approximation, minifloat approximation, and multiplier-free arithmetic are compared in the thesis work of Gysel[45]. An approximation framework, called Ristretto, was proposed with open-source code. TensorFlow also has 8 bit quantization support.

### 4.2.3 Network binarization and XNOR-Networks

An extreme case of quantization is binarization, i.e, the weights are reduced to either -1 or +1. The convolution computation basically turns into summation and subtraction. In XNOR network, both the filter and input to convolution layers are binary [46]. Compared with traditional 32 bit resolution computation, this results in a factor of 32 in memory savings and 58 times faster convolution operation. When evaluated on ImageNet classification, the binary weight network version of AlexNet achieved the same accuracy as the full-precision AlexNet. The authors also made the code available.

### 4.3 Software and Hardware Co-design

The previous subsection has shown great promise of algorithmic advances for mobile deep learning SWAP improvement. Even greater improvements can be achieved by taking an algorithmic and hardware co-design approach to make ASIC chips for mobile deep learning. For example, recent ASIC co-design work by Han, Liu, etc. has shown over four and three orders of magnitude improvement in energy efficiency than CPU and GPU, respectively. It also achieved over two and one order of magnitude improvement in speed than CPU and GPU, respectively on nine DNN benchmarks [47].

### 4.4 Improved Mobile Deep Learning System Performance using Domain Specific Data

The recent advances in deep machine learning algorithms have enabled many useful mobile deep learning applications developed for real use. However, for deep machine learning, the data can be just as important as the algorithm. It is not uncommon that the effort spent for collecting the right data is more rewarding than effort spent on improving the algorithm. The following subsection give some guidance on what to do with domain specific data.

Although there is a huge amount of publicly available datasets, mobile deep learning application may operate in different environments, use different sensors, or have different specifications. It is always a good idea to collect some real data from the kind of environment that the mobile deep learning application will be deployed. This data will be invaluable for improving the model or verifying the application performance. However, it is not always the best approach to build the model from scratch using new data. For mobile deep learning, there is less control of the environment, thus the testing data can have great variability. This means that to have a good performing model, a huge training dataset may be necessary. Unfortunately, data collection and data labelling in particular, can be very time consuming. Taking all these into consideration, the typical steps to proceed are:
1. Searching for pre-trained models similar to the application. For typical AI applications in computer vision, speech recognition, and NLP, there are many high quality models already made publicly available. Please refer to section 5.4 for a detailed list of existing AI application models.
2. Fine tune pre-trained models on domain specific data.
3. Leverage existing frameworks to deploy application.

Making use of existing models and frameworks can save a great deal of effort. However, deciding which part of the pre-trained model to be fine-tuned can be a work of art. This depends on the amount of domain specific data acquired and the similarity between this data and the original data used to build the pre-trained model. A general guide line is given in the Table 3 [4].

**Table 3. General Guidance of How to Improve Domain Data Performance Leveraging Existing Out of Domain Model**

| Domain data size | Similarity to pre-trained model data | How to improve? |
|---|---|---|
| Large | High | Fine tune the N/last layer |
| Small | High | Fine tuning may overfit the model, Train a linear classifier on (N-1) layer activation |
| Small | Low | Train a linear classifier on lower layer activation, Higher layers are more data specific to original data |
| Large | Low | Train DNN from scratch using domain specific data |

As mentioned in this section, there is a great deal of progress made in deep learning algorithms and models for mobile applications. The next section gives more detailed list of resource available in publication, libraries, tool, data, and models.

## 5. MOBILE DEEP LEARNING RESOURCES FOR PRACTITIONERS

### 5.1 Computer Languages for Mobile Deep Learning

Some of the most popular computer languages used in machine learning include Python, Matlab, C++, Java, R, and CUDA. Different approaches to implement mobile deep learning requires different platform, tools, and programming languages. These languages offer different development efficiencies and actual computation efficiencies. Thus, different approaches may be taken at different stages of the development. In addition, to deploy the developed system on different hardware platform, additional computer language skills are required. For example, Android and iOS platforms use different computer languages. There are also tools/computer language developed to offer solutions applicable to multiple platforms.

### 5.2 Open source code base for Deep Learning on Mobile Devices

The field of mobile deep learning has a relative low barrier to entry as most researchers make their source code freely available online. The most popular website to host the source code is github. For example, SqueezNet and model compression work mentioned in previous subsections are available at https://github.com/DeepScale/SqueezeNet and https://github.com/topics/model-compression. Facebook AI Research has recently open sourced Detectron platform for object detection. The trained models can be deployed on the mobile devices using Caffe2 runtime. The Detectron is available at https://github.com/facebookresearch/Detectron. In other cases, multiple versions of the same algorithm based on different platforms or libraries are made available by the machine learning community. In these case, it is usually a good idea to compare ratings, and how well documented and maintained the algorithms are.

### 5.3 Deep learning model zoo

As mentioned in the previous section, there are many deep learning models made publicly available. These model offer a good starting point to enable you to develop mobile deep learning applications. A few of these model zoos are listed below.

- TensorFlow/Keras model zoo. This is arguably the largest model zoo. It contain two categories, one for TensorFlow official models, and one for TensorFlow research models. The official models use TensorFlow's high-level API and tends to be better maintained and tested to keep up to date with latest TensorFlow updates. The official models are very broad. For example, the Keras version of the official model contains the following DNN architectures: Xception, Vgg, ResNet, InceptionResNet, MobileNet, DenseNet, and NASNet. The research models are implementation of some of the latest DNNs using TensorFlow. Some examples are: attention models, generative adversary models, deep speech, and transformer. They are maintained by individual researchers. The model link is https://github.com/TensorFlow/models

- Microsoft cognitive toolkit. This toolkit provide over 60 models, covering application domains of speech, image, text, finance, time series, and reinforcement learning. The link is https://www.microsoft.com/en-us/cognitive-toolkit/features/model-gallery/

- Caffe/Caffe2 model zoo. The Caffe pre-trained models cover dozens of the most influential models in the field of image processing and computer vision, such as CNN, VGG, ResNet, etc. The link is https://github.com/BVLC/caffe/wiki/Model-Zoo and https://github.com/caffe2/models/

- MXNet model zoo. In additional to some of most popular models also covered by other zoos, this zoo contains network in network model, single shot detection model, LocationNet, movie rating model, and video game emulator. The link is https://mxnet.incubator.apache.org/model_zoo/index.html

To fit a model downloaded from the zoo to a specific mobile platform, some of the previously mentioned technologies should be applied to optimize its SWAP metrics. Its performance should be benchmarked for mobile application.

### 5.4 Benchmarking Mobile Deep Learning Platform Performance

Modern Smartphones are capable of providing a computer power on the order of over 10 billion FLOPs. The benchmark of mobile deep learning should consider metrics such as accuracy, model size, and speed/execution time. . One benchmark is to compare various deep learning models for image classification on iPhone 7 [6][7]. The performance is summarized in Table 4.

**Table 4. Benchmark Image Classification Model Performance on iPhone 7 [6]**

| Model Architecture | Top1 Acc (%) | Model Size (MB) | Execution time (ms) |
|---|---|---|---|
| vgg16 | 71 | 553 | 208 |
| inception v3 | 78 | 95 | 90 |
| ResNet 50 | 75 | 103 | 64 |
| mobilenet | 71 | 17 | 32 |
| sqeezenet | 57 | 5 | 24 |

The second one is for AI benchmark on four chipsets (Qualcomm, HiSillicon, MediaTck, and Samsung) on Android smartphones [8]. The paper compared eight tasks in deep learning as AI benchmarks, including image classification, face recognition, deblurring, super-resolution, segmentation, and enhancement. Results are obtained with this benchmark from over 10,000 mobile devices and more than 50 different mobile SoCs. Detailed comparisons are reported at http://ai-benchmark.com/ . Some of the conclusions are:

- The easiest way to using deep learning on Android is to use TensorFlow Mobile framework
- TensorFlow Lite is an option, but we recommended for tasks more complicated than image classification
- For specific device or SoC, proprietary SDK can be used, but it is not as easy and convenient
- Caffe2 and other frameworks are much less popular with almost no tutorial and problem descriptions
- The applicability of model quantization is limited, performance may not be reliable

## 6. MOBILE DEEP LEARNING APPLICATIONS

### 6.1 Robotics

Robotics is being used increasingly in manufacturing industries, warehouse, agriculture, human access denied areas, and health care. Its use in the home is mostly limited to floor cleaning. As the AI technology progresses, robotics has many roles to play in the near future as envisioned by many scientists and engineers. In addition to the speech, NLP, and computer vision tasks mentioned in previous sections, robotics faces a few application specific tasks of learning, reasoning, and embodiment. Some recent reviews of the state of art and discussion of its potential and limitations are available [9] [10]. One key technique needed to make progress is to enable robots to autonomously acquire skills from sensory data. The robots need to be able to perform a general task without given specific instruction. This is also called learning to learn.

### 6.2 Autonomous driving

The advantages of Autonomous driving are numerous and can bring true revolution to transportation and life style in general. Autonomous driving is very popular in the industry as more capable companies are doing road testing of their own vehicles. Deep learning has revolutionized the perception modules for autonomous driving. Usually the perception is achieved by multi-modal sensors (e.g. EO camera, LiDARs, and Radras). Therefore, how to fuse multiple sensors in deep perception pipeline and achieve a nice model accuracy and computational cost trade-off is a very important and yet open topic. A very good discussion on this multi-modal fusion topic can be find here [53]. Many other interesting papers on applications of deep learning for autonomous driving have been published recently [11][12][13]. An empirical evaluation of CNN on highway driving has shown its capabilities of land and vehicle detection in real time [11][13]. A MultiNet has been proposed for real time solving of joint classification, detection, and semantic segmentation via a unified architecture [14]. Autonomous driving is another very important application of mobile deep learning that has a high hopes to become reality in the very near future. The obstacle to autonomous driving appears to be in driving in extreme conditions, in addition to non-technical issues such as ethical and legal.

### 6.3 Healthcare

While the health care resources may be limited, smartphone devices are increasingly available globally.. Applications of deep learning to improve the healthcare services has a potentially profound impact on well-being and the economy. With so many built-in sensors, Smartphones can be used as "medical" devices to monitor the physical and psychological status of individuals. For example, detection of depression is possible using audio and text in conversation [21]. In addition, the camera and speaker can serve as eyes and voice for people with vision and speech disorders. A special issue of AI for mobile health data analysis and processing to cover these topics are contained in [15][16][17].

### 6.4 Biometrics

Similar to many other fields of pattern recognition problems, deep learning is the leading solution for many biometric recognition modalities, including face, voice, keystroke, fingerprint, finger vein, iris, and gesture recognition [18][19][20]. As the smartphone contains so much user-private data, security is of high importance. Most smartphones have built-in fingerprint and or face recognition biometrics. As we can envision, for future shareable autonomous driving vehicles, the mobile biometrics component should also be a must for security purposes. To some extent, biometrics is an almost solved problem for cooperative subjects in close distance. One of key unsolved problem is for uncooperative subjects at a long standoff distance [18].

**6.5 Personal Assistance, Mobile Multi-Media, Augmented Reality, and Entertainment**

Personal assistance, mobile multi-media and people interaction are arguably the most applied domains of mobile deep learning technologies [22]. A recent report suggests over 47 million U.S. adults already have access to smart speaker, not to mention the personal assistance function available on the smartphone. Face detection is one well-known function on mobile phones. AI is also making an impact on the entertainment industry. AI is capable of generating text, imagery, and possibly video in the future. Entertainment content delivery, user engagement, and augmented reality can all be improved with AI technologies [23][24].

**6.6 Defense**

As AI technology progresses, it opens up applications to the defense industry. Semi-autonomous UAVs may be developed using some technologies similar to those for autonomous driving. Many of the computer vision technologies can be adapted for automatic target detection, recognition, and tracking. Deep learning is applicable to the full spectrum of electronic warfare sensing. Most of technologies developed for civil use of robotics may be applicable to the defense domain for transportation of military supplies and other tasks. Regardless of the technology readiness level, scientists, engineers, media, and politicians should all be very careful when promoting AI for autonomous weapons [25]. Many ethics, legal, regulatory, and policy issues are not yet resolved.

## 7. MOBILE DEEP LEARNING CHALLENGES AND FUTURE WORK

**7.1 Automatic Deep Machine Learning for mobile devices**

As mentioned in previous sections, a deep learning for mobile device application has to be optimized for SWAP. There are many hierarchical parameters in the deep learning system that can be optimized. These are called hyper-parameter. Optimization of hyper-parameter is an active topic called automatic machine learning [51]. Some technology of automatic ML may be applied to mobile applications. The object function for mobile DL is more than just accuracy and can be a weighted as a function of accuracy, memory footprint, computing speed, and power consumption.

**7.2 Low quality data and ML for signal processing**

Mobile application often means the application is less constrained. Unlike the well-controlled data collected in the laboratory, mobile data is often noisier with more confounding factors. This could also mean some, if not a significant portion, of the labelled data can be wrong. Although DNN is robust to noisy label when the training data is large, it can be a problem with less training data [52]. The mobile deep learning engineer thus needs to be more careful to prepare the data for model training. Sometimes custom signal processing may be necessary to deal with some special phenomenon.

**7.3 Small training data, few shot learning, and transfer learning**

Mobile applications may often have less training data to start with, although unlabeled data may be more abundant. To kick start the system, few shot learning or transfer learning may be applied [48][49]. As more data becomes available, the initial model can be improved iteratively.

### 7.4 Evolving environment, online adaptation, and lifelong learning

Another more challenging problem for mobile deep learning is that the environment can change over time. In some extreme cases, the system may have to face some unforeseen conditions. These situations require the mobile system to be able to perform online adaption to the change of the data or capable of lifelong learning [50].

### 7.5 Privacy in Mobile Deep Learning

As some mobile deep learning directly works on user personal data, privacy is a concern. This is particularly true if the data are stored and processed in the cloud. A hybrid deep learning architecture is recently proposed for privacy-preserving mobile analytics [26].

## 8. ACKNOWLEDGEMENT

The author would like to thank Dr. Stephen DelMarco for the invitation to submit and many constructive comments on this paper. The author would also like to thank Mrs. Jennifer Krischer for reviewing this paper.

## REFERENCES

[1] Cao, Q., "Embedded and mobile deep learning research notes", https://github.com/EMDL/awesome-emdl
[2] BLAS, https://en.wikipedia.org/wiki/Basic_Linear_Algebra_Subprograms
[3] Han, S., Liu, X., Mao, H., Pu, J., Pedram, A., Horowitz, M. A., & Dally, W. J. (2016, June). EIE: efficient inference engine on compressed deep neural network. In *Computer Architecture (ISCA), 2016 ACM/IEEE 43rd Annual International Symposium on* (pp. 243-254). IEEE.
[4] Iandola, F. N., Han, S., Moskewicz, M. W., Ashraf, K., Dally, W. J., & Keutzer, K. (2016). Squeezenet: Alexnet-level accuracy with 50x fewer parameters and< 0.5 mb model size. *arXiv preprint arXiv:1602.07360*.
[5] Rastegari, M., Ordonez, V., Redmon, J., & Farhadi, A. (2016, October). Xnor-net: Imagenet classification using binary convolutional neural networks. In *European Conference on Computer Vision* (pp. 525-542). Springer, Cham.
[6] "Squeezing Deep Learning Into Mobile Phones", Anirudh Koul, https://www.slideshare.net/anirudhkoul/squeezing-deep-learning-into-mobile-phones/
[7] Canziani, A., Paszke, A., & Culurciello, E. (2016). An analysis of deep neural network models for practical applications. *arXiv preprint arXiv:1605.07678*.
[8] Ignatov, A., Timofte, R., Szczepaniak, P., Chou, W., Wang, K., Wu, M., & Van Gool, L. (2018). Ai benchmark: Running deep neural networks on android smartphones. *arXiv preprint arXiv:1810.01109*.
[9] Harry A. Pierson & Michael S. Gashler (2017) Deep learning in robotics: a review of recent research, Advanced Robotics, 31:16, 821-835, DOI: 10.1080/01691864.2017.1365009
[10] Sünderhauf, N., Brock, O., Scheirer, W., Hadsell, R., Fox, D., Leitner, J., & Corke, P. (2018). The limits and potentials of deep learning for robotics. *The International Journal of Robotics Research*, *37*(4-5), 405-420.
[11] Huval, B., Wang, T., Tandon, S., Kiske, J., Song, W., Pazhayampallil, J., & Mujica, F. (2015). An empirical evaluation of deep learning on highway driving. *arXiv preprint arXiv:1504.01716*.
[12] Chen, C., Seff, A., Kornhauser, A., & Xiao, J. (2015). Deepdriving: Learning affordance for direct perception in autonomous driving. In *Proceedings of the IEEE International Conference on Computer Vision* (pp. 2722-2730).
[13] Bojarski, M., Del Testa, D., Dworakowski, D., Firner, B., Flepp, B., Goyal, P.,& Zhang, X. (2016). End to end learning for self-driving cars. *arXiv preprint arXiv:1604.07316*.
[14] Teichmann, M., Weber, M., Zoellner, M., Cipolla, R., & Urtasun, R. (2018, June). Multinet: Real-time joint semantic reasoning for autonomous driving. In *2018 IEEE Intelligent Vehicles Symposium (IV)* (pp. 1013-1020). IEEE.
[15] Artificial Intelligence to Prevent Mobile Heart Failure Patients Decompensation in Real Time: Monitoring-Based Predictive Model, Nekane Larburu, Arkaitz Artetxe, Vanessa Escolar, Ainara Lozano, and Jon Kerexeta Research Article (11 pages), Article ID 1546210, Volume 2018 (2018)

[16] User Evaluation of the Smartphone Screen Reader Voiceover with Visually Disabled Participants, Berglind F. Smaradottir, Jarle A. Håland, and Santiago G. Martinez Research Article (9 pages), Article ID 6941631, Volume 2018 (2018)
[17] Wearable DL: Wearable Internet-of-Things and Deep Learning for Big Data Analytics—Concept, Literature, and Future, Aras R. Dargazany, Paolo Stegagno, and Kunal Mankodiya Review Article (20 pages), Article ID 8125126, Volume 2018 (2018)
[18] Y. Deng, "Advances in Mobile and Remote Biometrics", Tutorial talk, IEEE Int. Conference on Biometrics Theory, Applications and Systems (BTAS), Sep 2016.
[19] Y. Zhong, Y. Deng, edit, "Recent Advances in User Authentication Using Keystroke Dynamics Biometrics", Science Gate Publishing, ISBN 978-618-81418-3-4, Jan, 2015.
[20] Bhanu, B., Kumar, A., "Deep Learning for Biometrics", Springer book, 2017, ISBN 978-3-319-61657-5.
[21] Alhanai, T., Ghassemi, M., James, G., "Detecting Depression with Audio/Text Sequence Modeling of Interviews", Interspeech 2018.
[22] Ota, K., Dao, M. S., Mezaris, V., & De Natale, F. G. (2017). Deep learning for mobile multimedia: A survey. *ACM Transactions on Multimedia Computing, Communications, and Applications (TOMM)*, *13*(3s), 34.
[23] Elkahky, A. M., Song, Y., & He, X. (2015, May). A multi-view deep learning approach for cross domain user modeling in recommendation systems. In Proceedings of the 24th International Conference on World Wide Web.
[24] Oberweger, M., Wohlhart, P., & Lepetit, V. (2015). Hands deep in deep learning for hand pose estimation. arXiv preprint arXiv:1502.06807.
[25] Scharre, P., Army of None: Autonomous Weapons and the Future of War, W.W. Norton & Company Publish, 2018.
[26] Osia, S. A., Shamsabadi, A. S., Taheri, A., Katevas, K., Sajadmanesh, S., Rabiee, H. R., & Haddadi, H. (2017). A hybrid deep learning architecture for privacy-preserving mobile analytics. arXiv preprint arXiv:1703.02952.
[27] Koul, A., Squeezing Deep Learning Into Mobile Phones - A Practitioner's guide, Slideshare presentation, March 2017.
[28] Chen, C., Yao, J., etc, A comprehensive, cross-framework solution to convert, visualize and diagnose deep neural network models. https://github.com/Microsoft/MMdnn
[29] LeCun, Y., Bengio, Y., & Hinton, G. (2015). Deep learning. *nature*, *521*(7553), 436.
[30] Goodfellow, I., Bengio, Y., Courville, A., & Bengio, Y. (2016). *Deep learning* (Vol. 1). Cambridge: MIT press.
[31] NVidia GPU for Autonomous Driving, https://www.nvidia.com/object/autonomous-cars.html
[32] A list of Graphics cards for notebook and smartphone GPU Cards. https://www.notebookcheck.net/Smartphone-Graphics-Cards-Benchmark-List.149363.0.html
[33] A list of GPU for android [https://forum.xda-developers.com/showthread.php?t=2232607].
[34] Singh, M. P., & Jain, M. K. (2014). Evolution of processor architecture in mobile phones. *International Journal of Computer Applications*, *90*(4).
[35] NVidia, C. U. D. A. (2007). Compute unified device architecture programming guide.
[36] Lacey, G., Taylor, G. W., & Areibi, S. (2016). Deep learning on fpgas: Past, present, and future. *arXiv preprint arXiv:1602.04283*.
[37] Wang, C., Gong, L., Yu, Q., Li, X., Xie, Y., & Zhou, X. (2017). DLAU: A scalable deep learning accelerator unit on FPGA. *IEEE Transactions on Computer-Aided Design of Integrated Circuits and Systems*, *36*(3), 513-517.
[38] Mead, C. (1990). Neuromorphic electronic systems. *Proceedings of the IEEE*, *78*(10), 1629-1636.
[39] Indiveri, G., Linares-Barranco, B., Hamilton, T. J., Van Schaik, A., Etienne-Cummings, R., Delbruck, T., & Schemmel, J. (2011). Neuromorphic silicon neuron circuits. *Frontiers in neuroscience*, *5*, 73.
[40] D. F. Bacon, R. Rabbah, and S. Shukla. Fpga programming for the masses. Communications of the ACM, 56(4):56–63, 2013.
[41] Deng, Y., Chakrabartty, S., & Cauwenberghs, G. (2004, July). Analog auditory perception model for robust speech recognition. In *Neural Networks, 2004. Proceedings. 2004 IEEE International Joint Conference on* (Vol. 3, pp. 1705-1709). IEEE.
[42] Koyanagi, M., Nakagawa, Y., Lee, K. W., Nakamura, T., Yamada, Y., Inamura, K., & Kurino, H. (2001, February). Neuromorphic vision chip fabricated using three-dimensional integration technology. In *Solid-State Circuits Conference, 2001. Digest of Technical Papers. ISSCC. 2001 IEEE International* (pp. 270-271). IEEE.
[43] Lane, N., Bhattacharya, S., Mathur, A., Forlivesi, C., & Kawsar, F. (2016, November). Dxtk: Enabling resource-efficient deep learning on mobile and embedded devices with the deepx toolkit. In *Proceedings of the 8th EAI International Conference on Mobile Computing, Applications and Services, ser. MobiCASE* (Vol. 16, pp. 98-107).

[44] Han, S., Mao, H., & Dally, W. J. (2015). Deep compression: Compressing deep neural networks with pruning, trained quantization and Huffman coding. *arXiv preprint arXiv:1510.00149*.

[45] "Ristretto: Hardware-Oriented Approximation of Convolutional Neural Networks", UC Davis Master Thesis of Philipp M. Gysel, 2016.

[46] Rastegari, M., Ordonez, V., Redmon, J., & Farhadi, A. (2016, October). Xnor-net: Imagenet classification using binary convolutional neural networks. In *European Conference on Computer Vision* (pp. 525-542). Springer, Cham.

[47] Han, S., Liu, X., Mao, H., Pu, J., Pedram, A., Horowitz, M. A., & Dally, W. J. (2016, June). EIE: efficient inference engine on compressed deep neural network. In *Computer Architecture (ISCA), 2016 ACM/IEEE 43rd Annual International Symposium on* (pp. 243-254). IEEE.

[48] Fawaz, H. I., Forestier, G., Weber, J., Idoumghar, L., & Muller, P. A. (2018). Transfer learning for time series classification. arXiv preprint arXiv:1811.01533.

[49] Fei-Fei, L., Fergus, R., & Perona, P. (2006). One-shot learning of object categories. *IEEE transactions on pattern analysis and machine intelligence*, *28*(4), 594-611.

[50] Silver, D. L., Yang, Q., & Li, L. (2013, March). Lifelong Machine Learning Systems: Beyond Learning Algorithms. In *AAAI Spring Symposium: Lifelong Machine Learning* (Vol. 13, p. 05).

[51] Feurer, M., Klein, A., Eggensperger, K., Springenberg, J., Blum, M., & Hutter, F. (2015). Efficient and robust automated machine learning. In *Advances in Neural Information Processing Systems* (pp. 2962-2970).

[52] Vahdat, A. (2017). Toward robustness against label noise in training deep discriminative neural networks. In *Advances in Neural Information Processing Systems* (pp. 5596-5605).

[53] Di Feng, Christian Haase-Schuetz, Lars Rosenbaum, Heinz Hertlein, Fabian Duffhauss, Claudius Glaeser, Werner Wiesbeck, Klaus Dietmayer, "Deep Multi-modal Object Detection and Semantic Segmentation for Autonomous Driving: Datasets, Methods, and Challenges", arXiv:1902.07830 , https://arxiv.org/pdf/1902.07830.pdf